\documentclass[a4paper,fleqn]{cas-dc}

\usepackage{amsmath}
\usepackage[colorlinks]{hyperref}
\usepackage[capitalise, noabbrev]{cleveref}

\usepackage[authoryear]{natbib}
\usepackage{amsmath,amsfonts}
\usepackage{algorithmic}
\usepackage{algorithm}
\usepackage{array}
\usepackage[caption=false,font=normalsize,labelfont=sf,textfont=sf]{subfig}
\usepackage{textcomp}
\usepackage{stfloats}
\usepackage{url}
\usepackage{verbatim}
\usepackage{graphicx}
\usepackage[utf8]{inputenc}
\usepackage{caption}
\usepackage{booktabs}
\usepackage{siunitx}
\usepackage{amssymb}
\usepackage{amsthm}
\usepackage{tikz}
\usepackage{pgfplots}
\usepackage{graphicx}
\usepackage{color}
\usepackage{subfloat}
\usepackage{verbatim}
\def\tsc#1{\csdef{#1}{\textsc{\lowercase{#1}}\xspace}}
\tsc{WGM}
\tsc{QE}

\begin{document}
\let\WriteBookmarks\relax
\def\floatpagepagefraction{1}
\def\textpagefraction{.001}

\shorttitle{<A Geometry-Aware Deep Network for Depth Estimation in Monocular Endoscopy>}    

\shortauthors{<Yang et al.>}  

\title [mode = title]{A Geometry-Aware Deep Network for Depth Estimation in Monocular Endoscopy}  



%
\author[1,2,3,4]{Yongming Yang}


\fnmark[1]

\ead{yangyongming@sia.cn}


\affiliation[1]{organization={Northeastern University},
            city={Shenyang},
            postcode={110819}, 
            country={China}}
\affiliation[2]{organization={State Key Laboratory of Robotics, Shenyang Institute of Automation, Chinese Academy of Sciences},
            city={Shenyang},
            postcode={110016}, 
            country={China}}
\affiliation[3]{organization={Institutes for Robotics and Intelligent Manufacturing, Chinese Academy of Sciences},
        city={Shenyang},
        postcode={110019}, 
        country={China}}
\affiliation[4]{organization={Key Laboratory of Minimally Invasive Surgical Robot, Liaoning Province},
        city={Shenyang},
        postcode={110016}, 
        country={China}}
        
\author[5]{Shuwei Shao}

\fnmark[1]

\ead{swshao@buaa.edu.cn}



\affiliation[5]{organization={Beihang University},
            city={Beijing},
            postcode={100191}, 
            country={China}}
\author[2,3,4]{Tao Yang}


\ead{yangtao@sia.cn}



\author[2,3,4]{Peng Wang}


\ead{wangpeng@sia.cn}

\author[6]{Zhuo Yang}


\ead{yangzhuocy@163.com}
\affiliation[6]{organization={General Hospital of Northern Theater Command},
            city={Shenyang},
            postcode={110016}, 
            country={China}}
\author[1]{Chengdong Wu}

\fnmark[*]

\ead{wuchengdong@mail.neu.edu.cn}

\author[2,3,4]{Hao Liu}

\fnmark[*]

\ead{liuhao@sia.cn}

\cortext[1]{Corresponding author: Hao Liu and Chengdong Wu}

\fntext[1]{Co-first authors}


\begin{abstract}
   Monocular depth estimation is critical for endoscopists to perform spatial perception and 3D navigation of surgical sites. However, most of the existing methods ignore the important geometric structural consistency, which inevitably leads to performance degradation and distortion of 3D reconstruction. To address this issue, we introduce a gradient loss to penalize edge fluctuations ambiguous around stepped edge structures and a normal loss to explicitly express the sensitivity to frequently small structures, and propose a geometric consistency loss to spreads the spatial information across the sample grids to constrain the global geometric anatomy structures. In addition, we develop a synthetic RGB-Depth dataset that captures the anatomical structures under reflections and illumination variations. The proposed method is extensively validated across different datasets and clinical images and achieves mean RMSE values of 0.066 (stomach), 0.029 (small intestine), and 0.139 (colon) on the EndoSLAM dataset. The generalizability of the proposed method achieves mean RMSE values of 12.604 (T1-L1), 9.930 (T2- L2), and 13.893 (T3-L3) on the ColonDepth dataset. The experimental results show that our method exceeds previous state-of-the-art competitors and generates more consistent depth maps and reasonable anatomical structures.  The quality of intraoperative 3D structure perception from endoscopic videos of the proposed method meets the accuracy requirements of video-CT registration algorithms for endoscopic navigation. The dataset and the source code will be available at \url{https://github.com/YYM-SIA/LINGMI-MR.}.
\end{abstract}




\maketitle

  \section{Introduction}
        
    Interventional endoscopy procedures, such as gastrointestinal endoscopy and colonoscopy, are effective tools for minimally-invasive diagnosis and treatment of abnormal conditions, where a key step is three-dimensional (3D) structure perception. The endoscopic augmented reality (AR) navigation system allows enhanced surgical visualization which requires registering pre-operative data (e.g., computed tomography (CT) scans) to the intra-operative endoscopic videos. The accuracy of video-CT registration algorithms primarily relies on the quality of intraoperative 3D structure perception from endoscopic videos. For reconstruction of 3D scenes, many methods use short-baseline stereo or structured light to acquire the desired binocular disparity or depth. Nevertheless, there is a growing need for spatially and temporally extended depth from monocular endoscopy due to the limited space within human anatomical cavities. Achieving this depth estimation goal will facilitate small-size monocular endoscopy that is more amenable to deployment with lower cost.
    
    Traditional methods have been developed by matching two-dimensional (2D) image pixels and their corresponding 3D spatial points in endoscopic scenes,~\textit{e.g.}, structure from motion (SfM) \citep{Leonard2018SfM}, and simultaneous localization and mapping (SLAM) \citep{chen2018slam}. To be specific, the feature-based methods \citep{mahmoud2017slam,turan2017non-rigid,song2018mis-slam,Kumar2019Jumping,Chen2019SLAM,lamarca2021defslam} perform feature matching and triangulation among multiple images using illumination and rotation-invariant feature descriptors. However, the overall scarce textures and the irregular camera motion in endoscopic scenes degrade the feature matching accuracy. 

    In recent years, learning-based monocular depth estimation methods have developed rapidly, aiming to infer depth from a single image. These methods can effectively inherit the key advantages of the conventional SfM and SLAM methods, and further solve the problem that the traditional feature-based methods fail for low-texture regions. For example, several supervised depth regression methods \citep{Visentini-Scarzanella2017bronchoscopy,mahmood2018CRF} use synthetic training data to compensate for the scarcity of high-accuracy RGB-Depth pairs in endoscopy. The depth difference between synthetic data and predicted data is used as a penalty term, which allows bypassing the appearance difference between adjacent frames and better dealing with the discontinuity of depth estimation caused by reflections and illumination variations. However, these methods ignore the important geometric structural consistency and are insensitive to sharply changing edges and small folds, inevitably resulting in the performance degradation and distortion of 3D reconstruction.

    Alternatively, the self-supervised methods \citep{Ma2019real-time,liu2020dense,liu2020Sinus,liu2020extremely,jamie2021Temporal,Recasens2021EDM,ozyoruk2021endoslam,shao2022AFM} make use of the disparity information contained in consecutive frames and minimize the appearance difference between the target and source frames to supervise the depth network. However, the appearance difference is often disturbed by reflections and illumination variations. There are two reasons for this confusion. On the one hand, the frequently-appearing inconsistent illumination breaks the inter-frame photometric consistency assumption. On the other hand, the endoscope and the light source are eye-in-hand tools that cause dramatic changes in the appearance of the same anatomical structure under endoscope motion, especially when the camera is close to the tissue surface. This may lead to ambiguous depth estimation and inaccurate anatomical reconstruction. 

    In contrast to the preceding methods, we pay attention to enforcing the geometric structural consistency for the entire scene. Concretely, we introduce gradient and normal losses to express the sensitivity to stepped edges and frequently fluctuations structures. Meanwhile, we propose a geometric consistency loss to constrain the global geometric surfaces structures. In addition, we develop a synthetic RGB-Depth dataset, which provides RGB images with reflection and illumination variations and the corresponding ground-truth depth maps, allowing for supervised learning applications. 
    
   \begin{figure*}[!htb]
    	\centering
    	\includegraphics[width=1\linewidth]{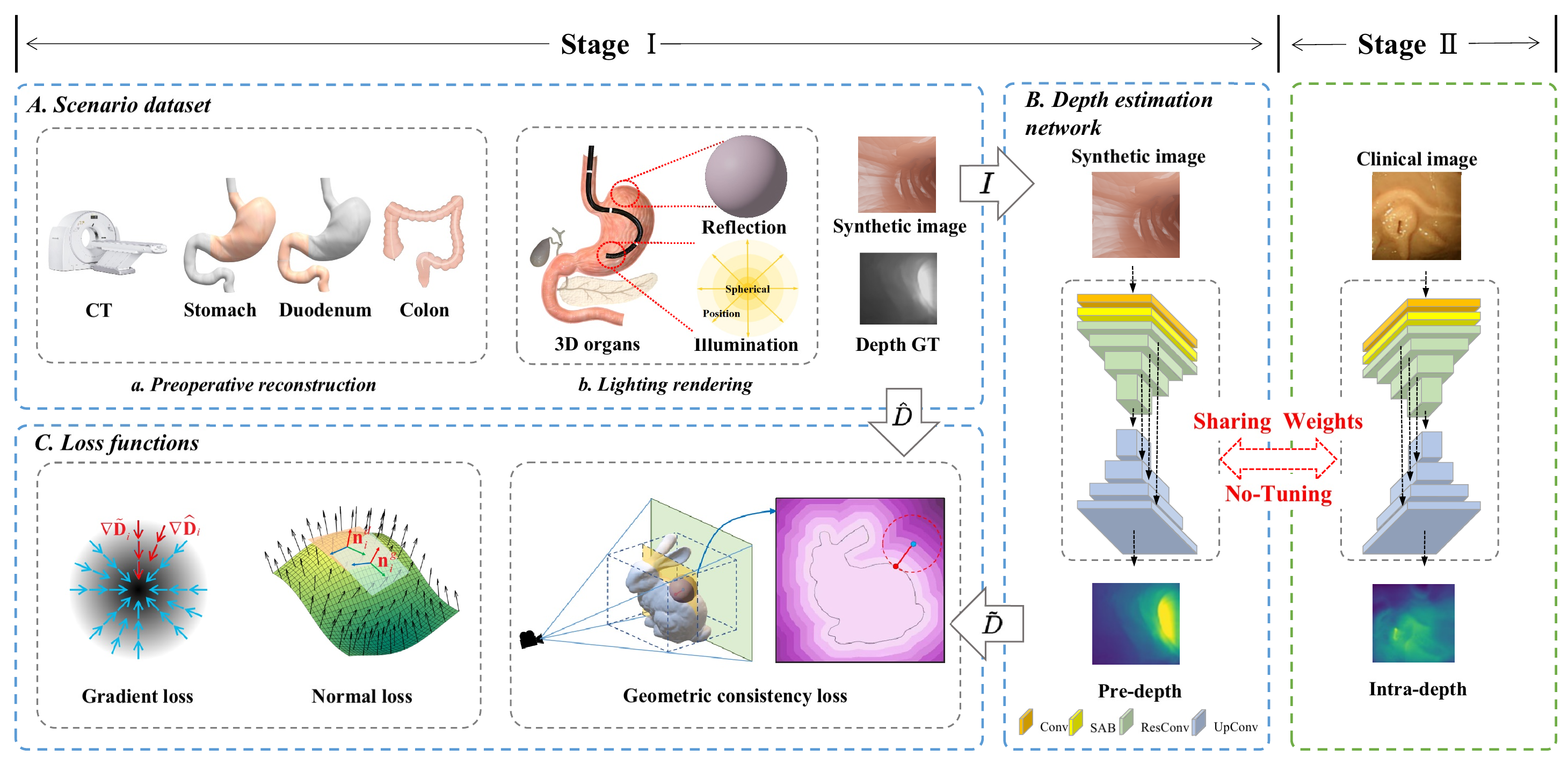}
    	\caption{The architecture of the proposed geometry-aware depth estimation system.}
    	\label{fig2:arch-of-system}
    \end{figure*}
    
    To summarize, the main contributions of this work are given as follows:
    \begin{itemize}
    	\item We introduce a synthetic RGB-Depth dataset tailored for interventional endoscopy. The dataset depicts the geometric structures under severe reflections and illumination variations. Besides, the dataset can reduce the cost of learning surgical skills in synthetic and real-world domains, and hence can essentially help with transferring skills learned in synthetic settings to real-world surgical scenarios.
    	\item We propose a novel geometry-aware depth estimation framework, which combines the strengths of the gradient and normal losses and geometric consistency loss. The framework enforces the geometric consistency constraints and boosts the reconstruction performance for stepped edges, small structures and global geometric surfaces structures.
    	\item Detailed experiments and analysis are conducted on the EndoSLAM dataset, Colondepth dataset and clinical images, indicating that the proposed method exceeds previous state-of-the-art competitors and generates more consistent depth maps and reasonable anatomical structures.
    \end{itemize}
    
     The rest of the paper is organized as follows: Section 2 describes the related works according to the classification of fully supervised, self-supervised and geometric priors for monocular depth estimation. In Section 3, geometry-aware depth estimation is described in detail. Section 4 describes experiments, details about the training parameter, quantitative comparison and generalization across different datasets and validation on clinical images. Finally, Section 5 discusses the limitations and future plans and offers some concluding remarks.  

    \begin{figure}[!htb]
    	\centering
    	\includegraphics[width=0.95\linewidth]{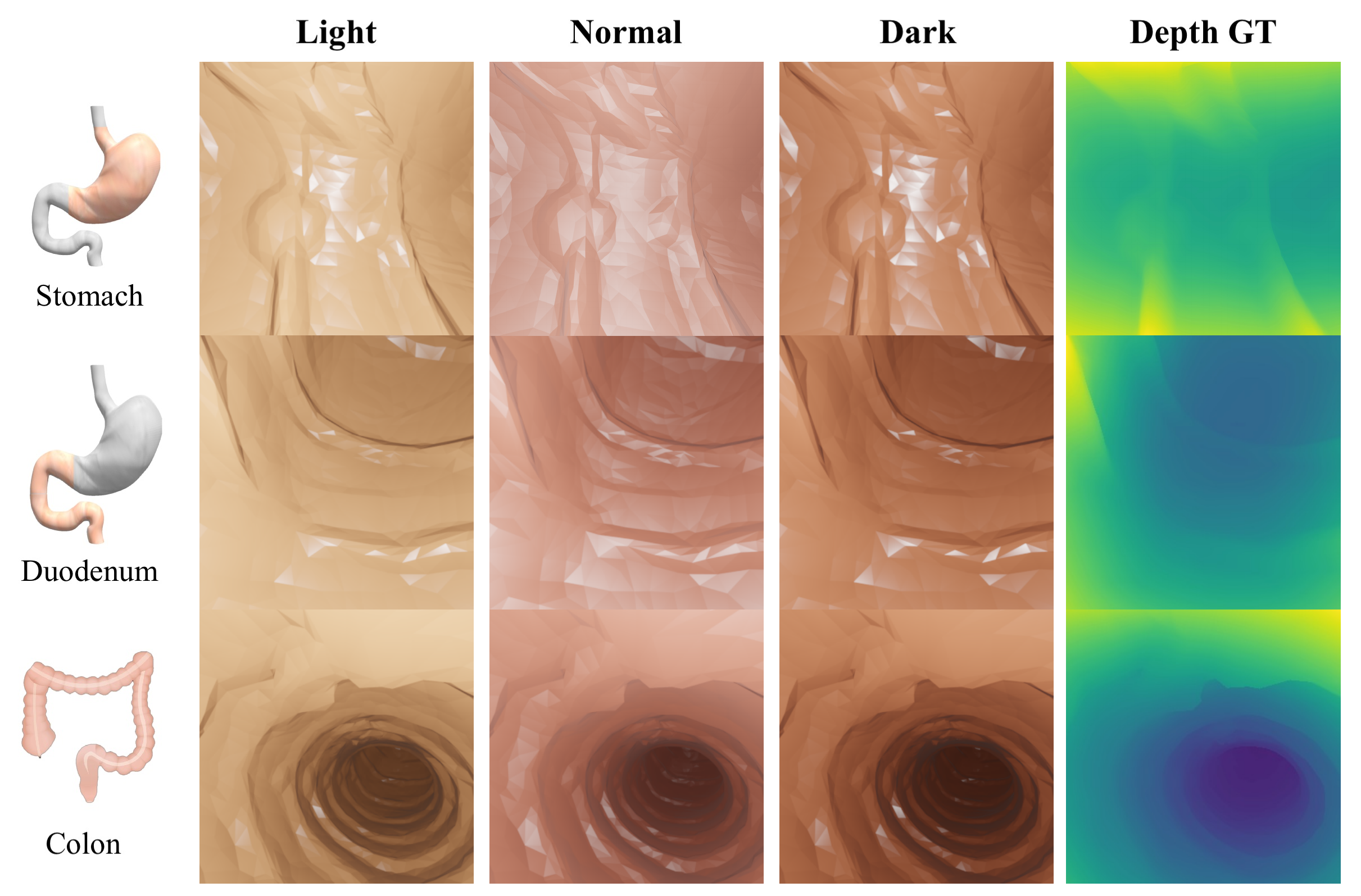}
    	\caption{Samples of the synthetic scenario dataset with different digestive parts and lighting conditions.}
    	\label{fig3:dataset}
    \end{figure}
	
    \section{Related Work}
      \subsection{Fully supervised monocular depth estimation}
    
    As a seminal work, \cite{eigen2014depth} proposed one of the first deep learning-based studies using a multi-scale network that combines global and local strategies for depth prediction. The coarse network is used to predict the overall trend, and the fine network is used to perform local optimization on the overall trend. Since then, numerous heuristic works have continued to improve the accuracy and usability of monocular depth estimation. \cite{Liu2014CNN} combined convolutional neural networks (CNN) with continuous conditional random fields (CRF) to estimate depth from a single image. \cite{laina2016FCN} proposed a residual  fully convolutional networks (FCN) architecture for monocular depth estimation, which has a deeper structure without post-processing. \cite{Cao2017FCRN} equaled the depth estimation as a pixel-level classification problem and trained a residual network for predicting the class corresponding to each pixel after discretizing the depth value. During the decoding stage, \cite{lee2019BTS} employed local-planar guidance layers rather than conventional upsampling layers to obtain full-resolution features. 
    However, it is challenging to use stereo and structured light cameras for ground truth acquisition of endoscopic scenes, and even if possible, these cameras may still not be able to obtain dense and accurate RGB-Depth dataset due to effects such as sparse textures and non-Lambertian reflections.     
    
    Several works based on simulation-to-real attempted to address the lack of supervised datasets by training synthetic dense depth maps generated from CT.  \cite{Visentini-Scarzanella2017bronchoscopy} used CT data and background-free endoscopic video simulations to train a fully-supervised learning architecture for depth estimation. Another transcoder network was then used to convert real endoscopic video frames into depth maps. Similarly, \cite{mahmood2018CRF} presented a supervised depth estimation approach where CRFs were combined with convolutional neural networks (CNN) for topography reconstruction. \cite{kai2021depth} used synthetic data to train a generative adversarial network (GAN) model and used real colonoscopy videos to tune the network by temporal consistency inter-frames. 
    Still, these methods ignore the important geometric structural consistency, resulting in degraded performance and distorted reconstructions of sharply varying edges and small folds.

      \subsection{Self-supervised monocular depth estimation}
    
    To reduce the production cost of supervised learning datasets, \cite{garg2016unsupervised} used left-right image pairs to estimate depth maps without depth labels, which works like an autoencoder. \cite{godard2017unsupervised} further improved this concept by exploiting the consistency of stereo views for unsupervised prediction. Inspired by the concept of temporal illumination consistency, later works have been extended with multi-scale appearance matching loss \citep{godard2019digging}, sequence information \citep{jamie2021Temporal} , repairing photometric loss \citep{vankadari2022sun}, moving instance instance loss \citep{yue2022self} and multilayer perception \citep{zheng2023self}. However, these self-supervised training methods typically rely on an assumption of the inter-frame constant brightness and material properties, which may not be applicable to interventional endoscopy procedures due to the endoscopic characteristics. 
   
    \begin{table}[!htb]
    	\centering
    	\caption{Dataset acquisition specifications.}
    	\includegraphics[width=0.9\linewidth]{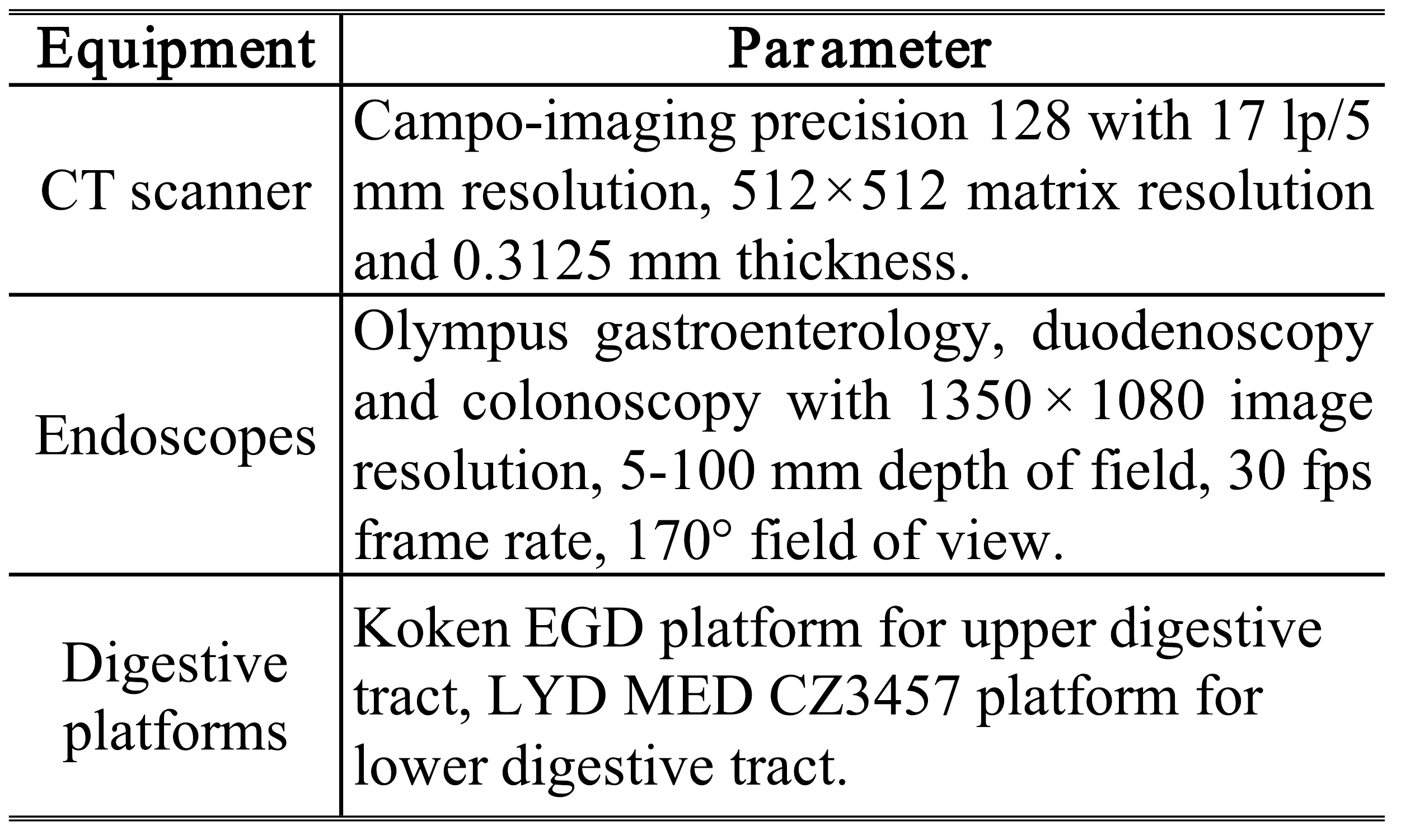}
            \label{table1:equipment}
    \end{table}
    
    Some exploratory works have been proposed for endoscopic self-supervised depth estimation. \cite{Ma2019real-time} achieved real-time dense reconstruction of colon images through deep-learning based dense SLAM. This technique effectively revealed hidden areas of the colon folds.  \cite{liu2020dense,liu2020Sinus,liu2020extremely} used video sequences (with corresponding SfM outcomes) and self-supervised learning strategies in order to train dense feature extraction and depth estimation modules. Using the Monodepth2 \citep{godard2019digging} network, \cite{Recasens2021EDM} proposed a keyframe-based photometric approach for endoscopic scene reconstruction with few-texture sequences. Several recent methods by \cite{ozyoruk2021endoslam} and \cite{shao2022AFM} used an affine brightness transformation and an appearance flow framework to account for brightness variations, respectively. However, these methods rely on the appearance properties of endoscopic images. The appearance properties are heavily susceptible to the low-texture regions and brightness changes. In contrast, the geometric properties of the 3D anatomy are more robust.

    \begin{figure*}[htp]
    	\centering
    	\includegraphics[width=0.87\linewidth]{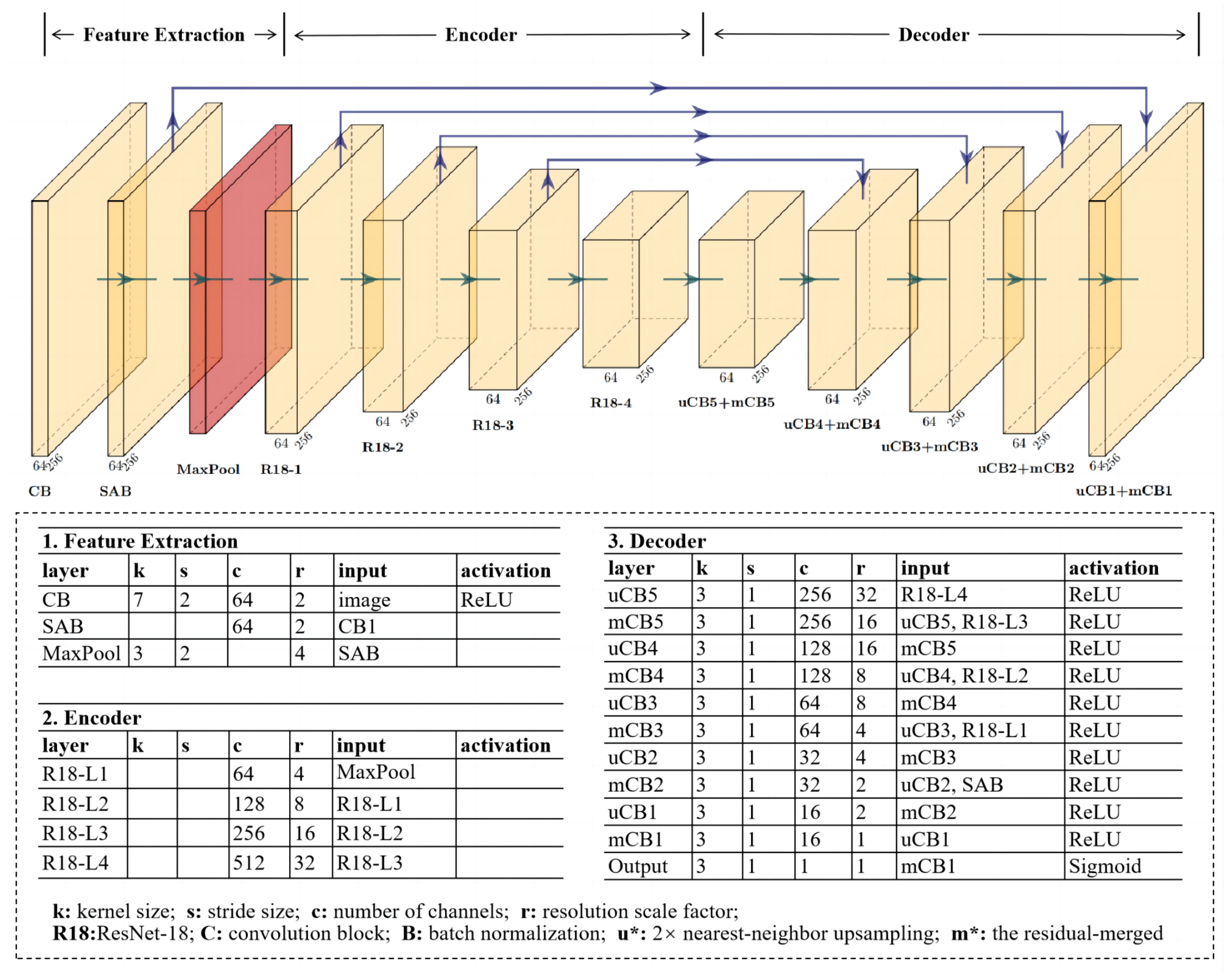}
            \caption{The overall architecture of the proposed depth estimation network.}
            \label{fig4:depth-net}
    \end{figure*}
  
      \subsection{Geometric priors for monocular depth estimation}

    Some works try to explicitly measure the inferred geometry by the geometric alignment loss, which are important for self-supervised depth learning. \cite{reza2018Geometric} estimated the point clouds consistency in consecutive sequences with iterative closest point (ICP) loss. It directly computed depth inconsistencies with the assumption that the scene is relatively static and the loss function can be approximated to be differentiable. \cite{fang2022Geometric} embedded the epipolar geometry into ICP to simplify the best-fit learning process in the self-supervised framework. Most recently, \cite{huang2022Geometric} enforced the geometric priors by leveraging Mask-ICP to compare stereo laparoscopic point clouds.    
    
    In contrast to the previously depth estimation methods for endoscopy, we introduce a novel geometry-aware loss of the whole scene to avoid discontinuous depth maps. In addition, we introduce an endoscopic RGB-Depth dataset with fully geometric variations, which allows supervised methods to exert better performance and generalization ability.

    \section{Geometry-aware depth estimation}
       
    In this section, we elaborate on the proposed geometry-aware depth estimation framework, as shown in \cref{fig2:arch-of-system}. Specifically, the framework contains two stages, where the stage I is divided into three parts: A. Scenario dataset, B. Depth estimation network and C. Loss functions, and the stage II is a zero-shot application of the trained depth network on clinical images.

     \subsection{Scenario dataset}

    We build a equipment to collect preoperative CT images and corresponding intra-operative endoscopy videos. The essential components include a high-resolution CT scanner for obtaining 3D measurements of the digestive tract, endoscopes and digestive platforms. The equipment specifications are given in \cref{table1:equipment} and samples of the synthetic data are shown in \cref{fig3:dataset}.
    
    We scan the digestive tract using a CT scanner, manually segment the CT images, and extracted surface meshes. Furthermore, we use Blender, an open-source 3D modeling and rendering tool, to simulate the endoscope imaging process, created via rendering synthetic RGB images with corresponding depth maps. A virtual camera with additional point lights can be programmed to follow the desired path in the image synthesis model. 
   
    To fully simulate the effects of real-world lighting conditions, we run the \begin{figure}[!htb]
    	\centering
    	\includegraphics[width=0.95\linewidth]{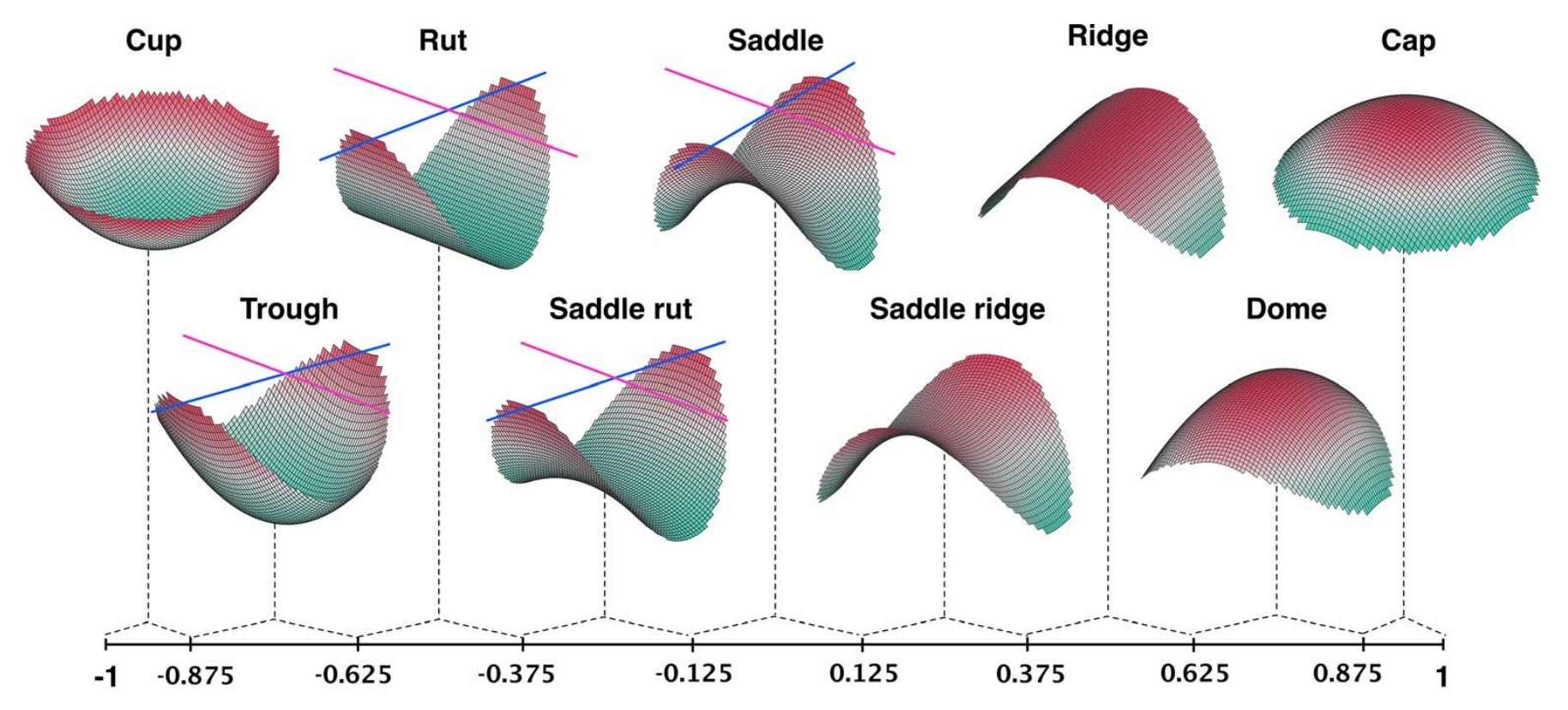}
    	\caption{Shape-index scale \citep{thakku2015shape-index} with corresponding shape.}
    	\label{fig5:shape}
    \end{figure} render in a loop through different paths, resulting in multiple data subsets with different texture, color and intensity settings. The texture samples are obtained from a real endoscopic image, and the associated color and intensity values are varied for each sample. We end up with over 10,000 RGB-Depth pairs with a minimum depth of 0.01 (in terms of the simulation unit) and a maximum depth of 100. Both the dataset and its creation module will be made publicly available.

    \subsection{Depth estimation network}
    
    Our depth estimation network mainly follows the design in Manydepth \citep{jamie2021Temporal} with some modifications, as shown in \cref{fig4:depth-net}. We use ResNet18 as the backbone for the encoder, and add a spatial attention block (SAB) \citep{ozyoruk2021endoslam} at the ResNet18 \citep{he2016deep} input to enhance the spatial features. As for the decoder, we use upsampling convolution layers followed by batch normalization and leaky ReLU nonlinear activation for each scale of skip-connected features from the encoder. Our network accepts a frame of 3-channel RGB image with a resolution of 320 × 320 pixels. The output is a scaled depth map with the same resolution as the input.

     \subsection{Loss functions}
    
    \textbf{Depth loss.} To supervise the depth network, we measure the difference between depth prediction and ground truth using the L1 distance,
       
    \begin{equation}
    	{{\mathcal{L}}_{depth}}={{\left\| \mathbf{\tilde{D}}_{i}^{{}}-{{\widehat{\mathbf{D}}}_{i}} \right\|}_{1}},
    \end{equation}where $\mathbf{\tilde{D}}_{i}^{{}}$ and ${\widehat{\mathbf{D}}}_{i}$ are the predicted and ground-truth depth maps, respectively.

    \textbf{Smoothness loss.} As in \citep{godard2017unsupervised}, we use a smoothness loss to encourage the smoothness property of the depth prediction,
    \begin{equation}    
    	{{\mathcal{L}}_{smooth}}={{({{e}^{-\nabla {{\mathbf{I}}_{i}}}}\cdot \nabla \mathbf{\tilde{D}}_{i}^{{}})}^{2}}.
    \end{equation}
    
    Although these two losses are sensitive to changes in depth direction, they are not particularly sensitive to changes in the imaging plane direction.

    \textbf{Gradient loss.} As shown in \cref{fig3:dataset}, we can observe from the synthetic scenario dataset that endoscopic images are composed of many 
        \begin{figure}[!htb]
    	\centering
    	\includegraphics[width=0.95\linewidth]{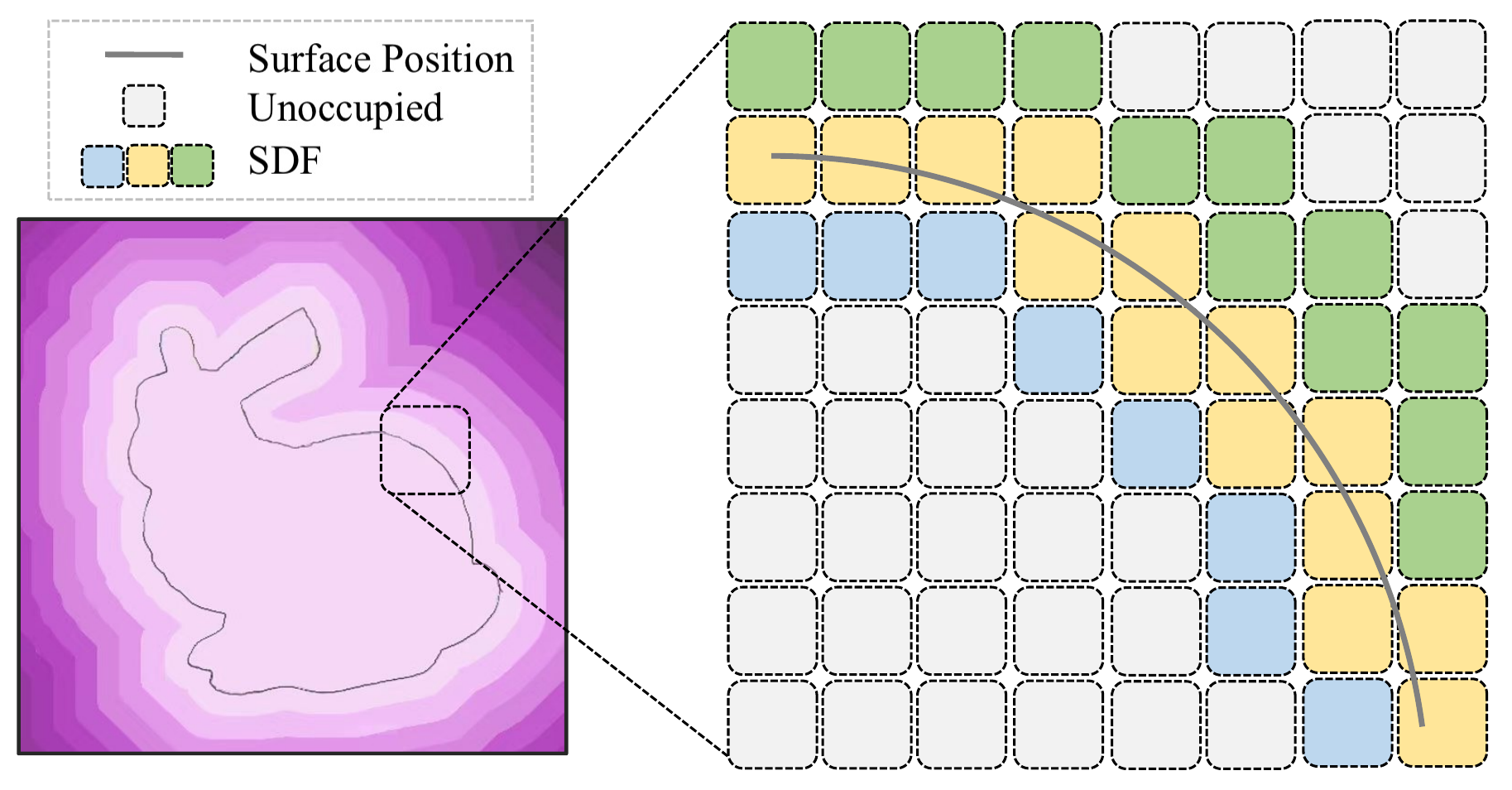}
    	\caption{
    		Signed distance field function.}
    	\label{fig6:sdf}
    \end{figure} stepped edge structures, and the sensitivity of edge distortion and blur is critical for recovering 3D structure. We investigate into the anatomical differences of stomach wall, duodenal papilla, diverticula and colon folds. Each of these anatomical tissues has some unique characteristics. Stomach wall and colon folds are ridge-shape features with lower curvature values, while duodenal papilla and diverticula are cap-shape structure with slightly higher curvature values. The different anatomical structures can be matched by the curvature value corresponding to the shape index \cref{fig5:shape}.

    Because the depth and smoothness losses are not sensitive to edge fluctuations, the stepped edge structures in depth maps trained with these two losses tend to be ambiguous. Thus, we introduce the following gradient loss function \citep{hu2019revisiting} to penalize such errors around stepped edge structures,
    \begin{equation}
    	{{\mathcal{L}}_{grad}}={{\left\| {{\nabla }_{x}}\mathbf{\tilde{D}}_{i}^{{}}-{{\nabla }_{x}}{{\widehat{\mathbf{D}}}_{i}} \right\|}_{1}}+{{\left\| {{\nabla }_{y}}\mathbf{\tilde{D}}_{i}^{{}}-{{\nabla }_{y}}{{\widehat{\mathbf{D}}}_{i}} \right\|}_{1}}.
    \end{equation}

    \textbf{Normal loss.} According to the results of prior depth estimation methods, depth maps may be generally approximated by a finite number of smooth surfaces with stepped edges. The gradient loss is sensitive to the changes in the imaging plane direction, and is complementary to the depth and smoothness losses. However, we have to select a modest weight on gradient loss because depth differences at such occluding edges can sometimes be large. In this case, the gradient loss does not penalize well for small structural errors like frequently fluctuations of surfaces. To obtain fine details in the depth maps and recover such small depth structures, we consider adding a normal loss \citep{hu2019revisiting} that measures the normal accuracy between the estimated depth map and the ground truth.
        
    We use the cosine-similarity loss for the surface normal generated from the depth map,
    \begin{equation}
    	{{\mathcal{L}}_{normal}}=1-\frac{\left\langle \mathbf{\hat{n}}_{i}^{{}},\mathbf{\tilde{n}}_{i}^{{}} \right\rangle }{{{\left\| \mathbf{\hat{n}}_{i}^{{}} \right\|}_{2}}\cdot {{\left\| \mathbf{\tilde{n}}_{i}^{{}} \right\|}_{2}}},
    \end{equation} where $ \mathbf{\tilde{n}}_{i}^{{}}={{\left[ -{{\nabla }_{x}}\mathbf{\tilde{D}}_{i}^{{}},-{{\nabla }_{y}}\mathbf{\tilde{D}}_{i}^{{}},1 \right]}^{T}} $, $ \mathbf{\hat{n}}_{i}^{{}}={{\left[ -{{\nabla }_{x}}{{\widehat{\mathbf{D}}}_{i}},-{{\nabla }_{y}}{{\widehat{\mathbf{D}}}_{i}},1 \right]}^{T}} $ are the surface normal associated with the predicted and ground-truth depth maps, respectively, and $\left\langle \cdot,\cdot  \right\rangle$ is the inner product operator.

    \textbf{Geometric consistency loss.} 
    \begin{figure*}[htb]
        \vspace{6pt}
        \centering
        \includegraphics[width=1.0\linewidth]{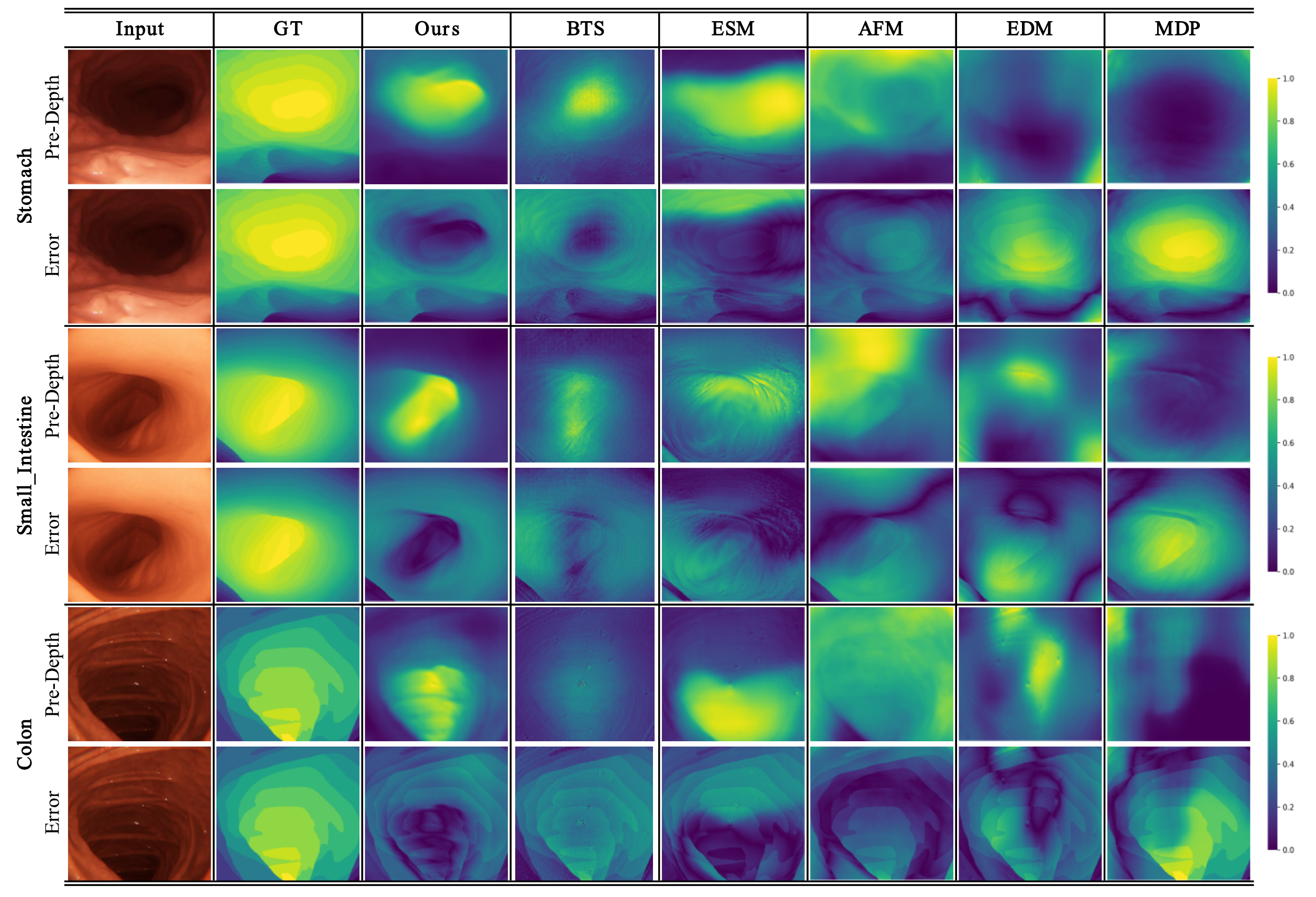}
        \caption{Qualitative comparison on the EndoSLAM dataset. The original input image, ground-truth depth, depth maps predicted, and error maps by our method, BTS, ESM, AFM, EDM, and MDP are shown from left to right, respectively.}
        \label{fig7:expr-a}
    \end{figure*} 
    In addition to being sensitive to stepped edges and small structures, the overall structure, such as a reasonable colon lumen structure and an accurate relative spatial relationship between gastric folds, is more important for endoscopic depth estimation. To constrain the global geometric surface structures, we introduce a geometric loss which is spread across all image pixels. This allows the geometric consistency constraints, and provides better supervision by all pixels during the depth estimation process. 
    
    Given a single image, our goal is to reconstruct the 3D geometry and textures of an anatomy while preserving the fine-grained details. However, due to the complex topology of the lumen structure details, it is difficult to express a geometry with an analytical formula. We only have some discrete observation points, such as a certain point inside the lumen structure, another point outside the lumen structure. To this end, we introduce implicit function which is a memory efficient representation for 3D surfaces. The details are described as follows:
    
    As shown in \cref{fig6:sdf}, a distance field (DF) stands for a function $\phi :{{\mathbb{R}}^{3}}\to \mathbb{R}$, which maps a 3D spatial point $\mathbf{q}$ to its minimum distance to a given surface $\mathcal{S}$, i.e.,
    \begin{equation}
    	\phi (\mathbf{q},\mathcal{S})={{\min }_{\mathbf{X}\in \mathcal{S}}}||\mathbf{X}-\mathbf{q}|{{|}_{2}},\mathbf{q}\in {{\mathbb{R}}^{3}},
    \end{equation}
    where $\mathbf{X}$ is an arbitrary point on $\mathcal{S}$ and $||\cdot||_{2}$ is the 2-norm of a vector. We take the signed distance field (SDF) 
    into account where the positive (resp. negative) SDF values are associated with points inside (resp. outside) the surface $\mathcal{S}$, i.e., ahead of (resp. behind) the surface from the camera viewpoint.

    To get the SDF of a depth map, we carry out back-projection from the image space. The projection mapping is denoted as,
    \begin{equation}
    	\pi :\mathbf{x}=\pi (\mathbf{X}),
    \end{equation}
    where $\mathbf{x}={{(u,v)}^{\top }}$ is an image point, and $\mathbf{X}={{(X,Y,Z)}^{\top }}$ is the corresponding 3D spatial point. 
    The inverse mapping of $\pi$ is
    \begin{equation}
    	{{\pi }^{-1}}:\mathbf{X}={{\pi }^{-1}}(\mathbf{x})={{\pi }^{-1}}(\mathbf{x},\mathcal{\tilde{D}}(\mathbf{x})).
    \end{equation} 

    Then, a surface back-projected from a depth map is acquired by 
    \begin{equation}
    	\mathcal{S}={{\pi }^{-1}}(\mathcal{I},\mathcal{\tilde{D}}),
    \end{equation}where $\mathcal{I}=\{\mathbf{x}\}$ is the image domain.

    \begin{table}[htb]
        \vspace{6pt}
        \centering
        \caption{Dataset and network parameter number.}
        \includegraphics[width=0.87\linewidth]{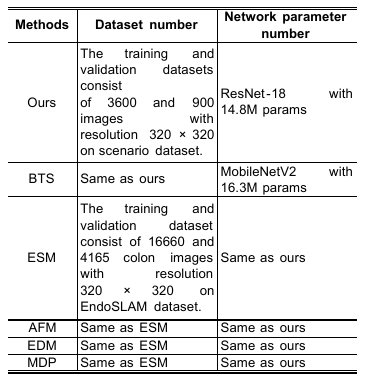}
        \label{Dataset and network params}
    \end{table}
    
    To evaluate the SDF loss, we consider a certain 3D space under the camera frame. Suppose ${{\mathbf{X}}_{lb}}$  and ${{\mathbf{X}}_{ub}}$  are the lower bound and the upper bound of the sample space of a surface, respectively. Each sample point is given by 
    \begin{equation}
    	\begin{matrix}
    		{{\mathbf{X}}_{ijk}}={{(\frac{i}{{{r}_{x}}-1},\frac{j}{{{r}_{y}}-1},\frac{k}{{{r}_{z}}-1})}^{\top }}*({{\mathbf{X}}_{ub}}-{{\mathbf{X}}_{lb}}),
      \\ i=0,...,{{r}_{x}},j=0,...,{{r}_{y}},k=0,...,{{r}_{z}}  \\
    	\end{matrix}
    \end{equation}where ${{r}_{x}},{{r}_{y}}$ , and ${{r}_{z}}$ denote the respective resolutions of the three dimensions, and $\ast$ is the element-wise vector multiplication operator. Then, the SDF loss is defined as the average inter-sample distance error:
    \begin{equation}
    	{{\mathcal{L}}_{sdf}}=\frac{\sum\limits_{k}{\sum\limits_{j}{\sum\limits_{i}{||}}}\phi ({{\mathbf{X}}_{ijk}},{{\pi }^{-1}}(\mathcal{I},\tilde{D})-\phi ({{\mathbf{X}}_{ijk}},{{\pi }^{-1}}(\mathcal{I},\hat{D}))|{{|}_{1}}}{{{r}_{x}}\cdot {{r}_{y}}\cdot {{r}_{z}}}.
    \end{equation}
    
    Apart from the explicit surface representation with 3D coordinates, the implicit SDF method spreads the spatial information across the sample grids, so that the error of each spatial point contributes to the total loss. In this way, the depth estimation network pays more attention to the global geometric surface structures rather than the local texture details only. We use the K-nearest neighbor (KNN) method implemented in Pytorch3D \citep{ravi2020pytorch3d} for differentiable minimization $\min (\cdot )$  of $\phi (\cdot )$  during network training.
    
    The overall depth loss is a weighted sum of the geometric consistency loss and additional pixel-based losses. This overall loss can be formulated as:
    \begin{equation}
    	\begin{aligned}
        {{\mathcal{L}}_{\text{total}}}={{\lambda}_{\text{1}}}({{\mathcal{L}}_{\text{de}pth}}+{{\mathcal{L}}_{smooth}})
        &\text{+}{{\lambda}_{\text{2}}}({{\mathcal{L}}_{grad}}+{{\mathcal{L}}_{normal}})\\
        &\text{+}{{\lambda }_{\text{3}}}{{\mathcal{L}}_{sdf}}
    	\end{aligned}
    \end{equation}where ${\lambda}_{\text{1}}$ denotes the weight of the depth and smoothness losses, ${\lambda}_{\text{2}}$ denotes the weight of gradient and normal losses, and ${\lambda}_{\text{3}}$ denotes the geometric consistency loss.
    
    \begin{figure}[htb]
    	\centering
    	\includegraphics[width=0.75\linewidth]{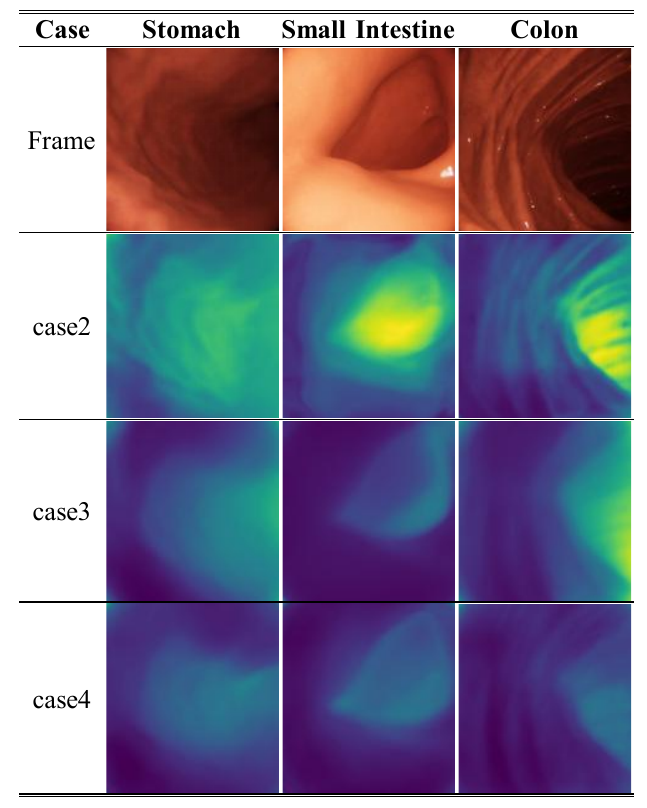}
        \caption{Visualization results of ablation study on the EndoSLAM dataset.}
        \label{Visualization-ablation study}
    \end{figure}
    
    Once the training process of the depth network is completed, we apply the trained depth network to real images with a zero-shot evaluation.

     \begin{figure*}[htb]
		\centering
		\vspace{6pt}
		\includegraphics[width=1.0\linewidth]{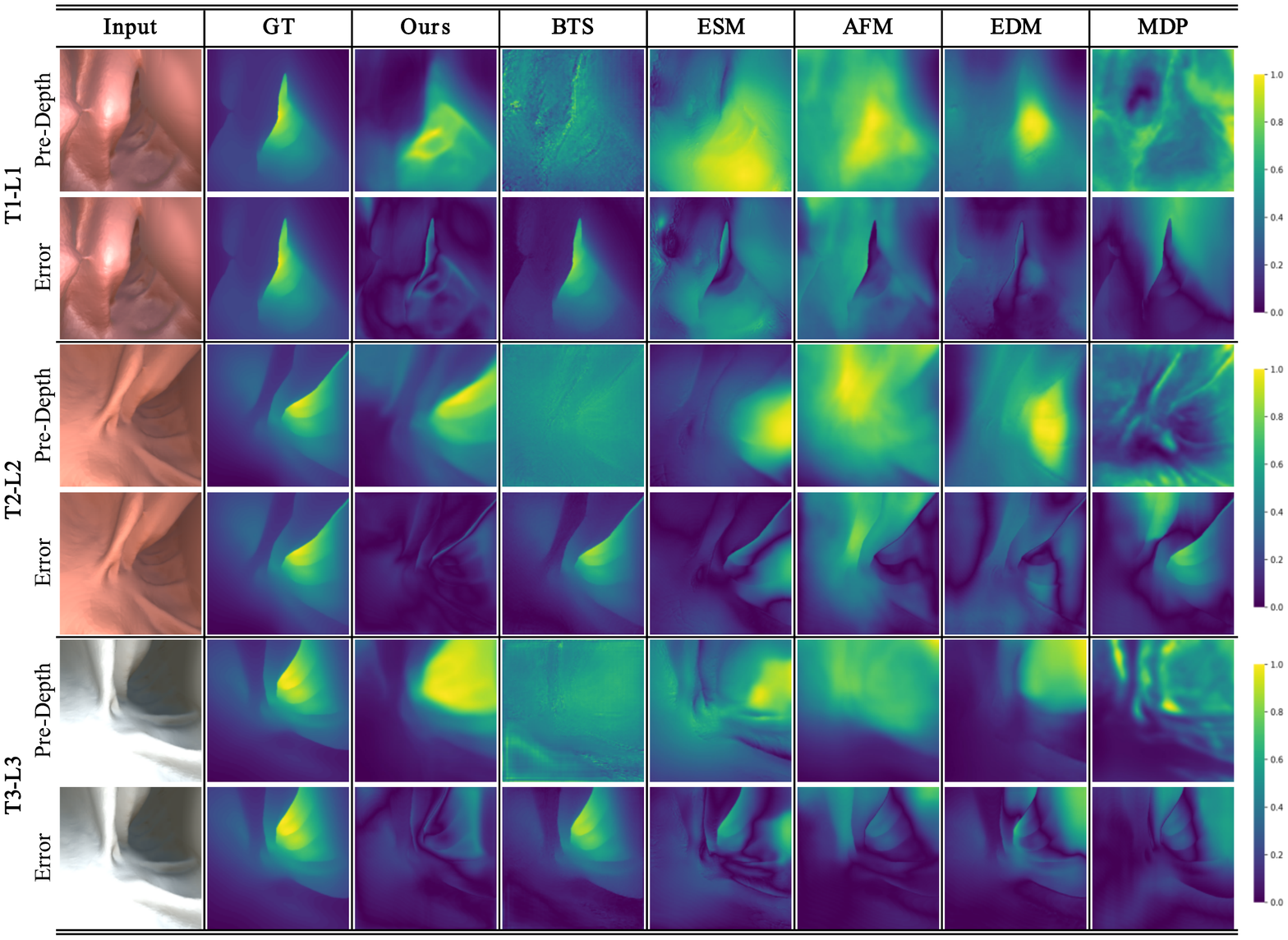}
		\caption{Generalization ability on the ColonDepth dataset. Three subsets are divided into groups with different material texture (T1, T2, T3) and lighting conditions(L1, L2, L3). Our method, BTS, ESM, AFM, EDM, and MDP are shown from left to right, respectively.}
		\label{fig8:expr-b} 
    \end{figure*}     
    
    \section{Experiments}
	In order to assess the depth estimation accuracy for the proposed framework and investigate different design considerations, we perform extensive experiments in this section.

	\subsection{Training parameter}
	
	The depth estimation network is trained in a fully-supervised manner on the introduced synthetic dataset. The training workstation has an Intel Core 3.6-GHz CPU with 8 threads, an Ubuntu 20.04 LTS operating system, and a NVIDIA RTX 2080TI GPU with 16 GB of RAM. We use the Adam optimizer with the following parameters: a weight decay of 0.01, weight parameters of $\beta_1$=0.9, $\beta_2$=0.999, a number of epochs of 18, a batch size of 20, and a learning rate of 0.0001 (1-18). We split the Scenario dataset into 3600, 900, and 300 frames for the training, validation and test sets, respectively. We conduct depth estimation on the EndoSLAM dataset, the ColonDepth dataset and clinical images. We select three sets of 1548 stomach frames, 1257 small-intestine frames, and 1062 colon frames to evaluate performance on the EndoSLAM dataset. In terms of the ColonDepth dataset, we conduct 3-fold cross validation with 364 T1-L1 frames, 364 T2-L2 frames, and 364 T3-L3 frames. Besides, we demonstrate quantitative point cloud results and qualitative surface reconstructions on the Kvasir dataset and clinical endoscopy videos. Compared with other competitive methods, our method does not use larger datasets and larger models as shown in \cref{Dataset and network params}. The evaluation datasets are described as follow:

	\textbf{EndoSLAM dataset} \citep{ozyoruk2021endoslam}. The EndoSLAM is captured from six porcine organs and includes CT scanning ground-truth data, clinical standard endoscope recordings of phantom colon and endoscopy videos, synthetically-generated data, and 3D point cloud data.
 
    \begin{table}[htb]
        \centering
        \caption{Ablation study on the EndoSLAM dataset.}
        \includegraphics[width=0.95\linewidth]{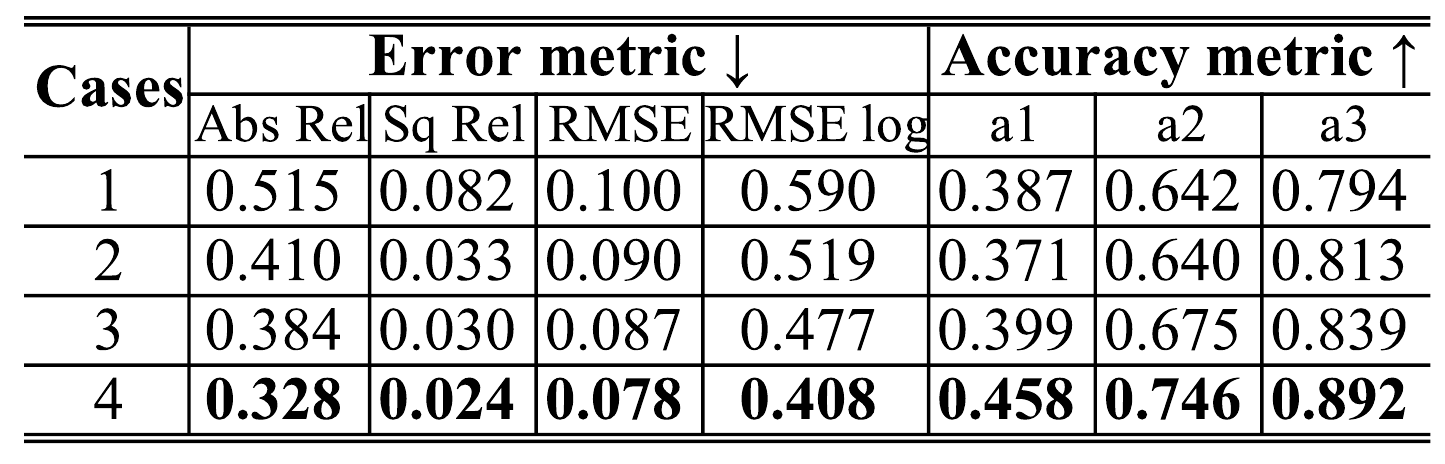}
        \label{table2:ablation}
    \end{table}

	\textbf{ColonDepth dataset} \citep{Rau2019colondepth}. The ColonDepth has more than 16,000 RGB images with corresponding depth maps. Each image is resized to 256×256 pixels. The data is clustered in groups based on texture and illumination patterns. By randomly translating and rotating the virtual camera, four to five different subsets are produced for each configuration.
	
	\textbf{Kvasir dataset} \citep{pogorelov2017kvasir}. The Kvasir consists of physician annotated and validated images showing anatomical landmarks and gastrointestinal endoscopic procedures. Image resolutions range from 720×576 to 1920×1072 pixels.

      \begin{table*}[htb]
		\vspace{6pt}
		\centering
		\caption{Quantitative results. ESM, AFM, EDM, MDP, BTS and our method are shown from left to right, respectively. The methods are benchmarked on the EndoSLAM dataset with 1548 stomach frames, 1257 small-intestine frames, and 1062 colon frames. The best results are in bolded, and the second best ones are underlined.}
		\includegraphics[width=\linewidth]{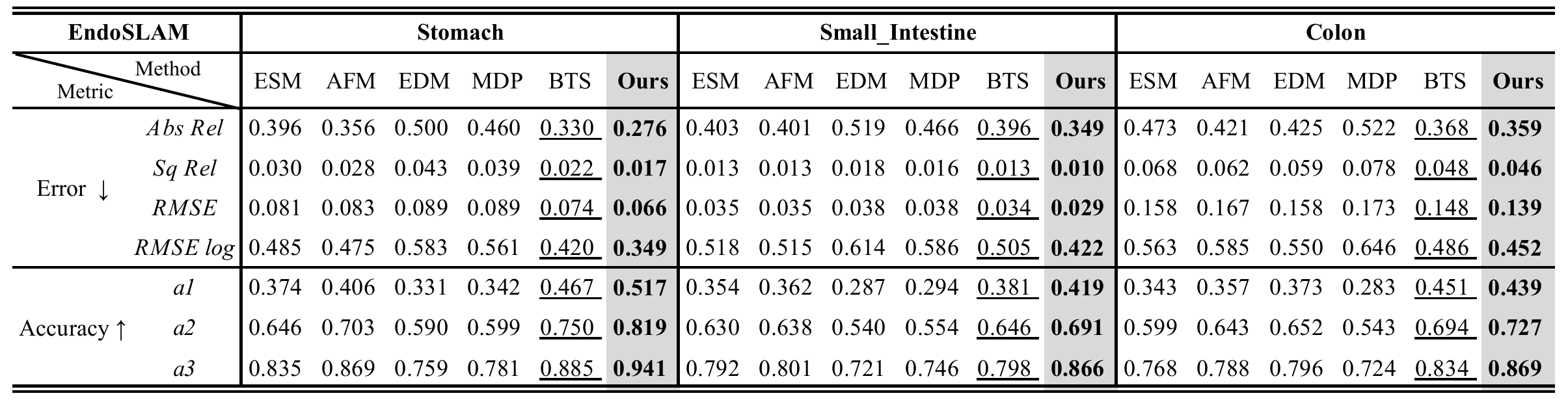}
		\label{table3:endoslam}
	\end{table*}
	
	\begin{table*}[htb]
		\centering
		\vspace{6pt}
		\caption{Generalization to the ColonDepth dataset. ESM, AFM, EDM, MDP ,BTS and our method are shown from left to right, respectively. The methods are benchmarked on the ColonDepth dataset with 364 T1-L1 frames, 364 T2-L2 frames, and 364 T3-L3 frames.}
		\includegraphics[width=\linewidth]{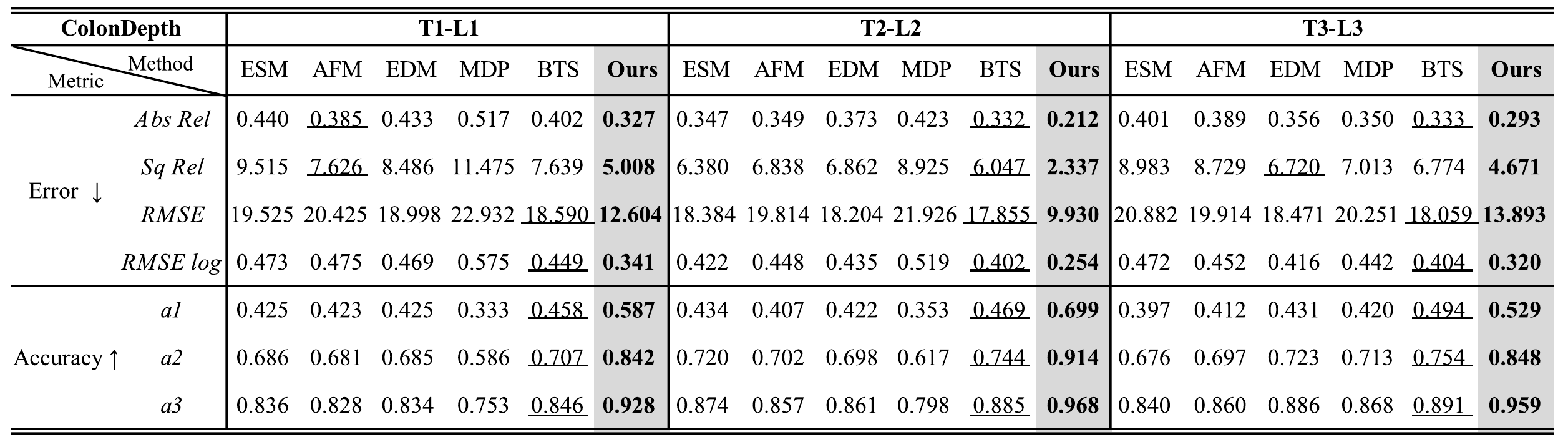}
		\label{table4:colondepth}
	\end{table*}

	 \subsection{Evaluation and discussion}
	
	The proposed method is thoroughly compared against the following state-of-the-art unsupervised approaches: Endo-SfM (abbr. ESM) \citep{ozyoruk2021endoslam}, AF-SfM (abbr. AFM) \citep{shao2022AFM}, Endo-Depth-and-Motion (abbr. EDM) \citep{Recasens2021EDM}, Monodepth2 (abbr. MDP) \citep{godard2019digging}. For fair unbiased results, all of the unsupervised methods are trained on the EndoSLAM dataset. The training and validation dataset consist of 16660 and 4165 colon images on UnityCam with resolution 320×320. To be more persuasive, we compare with a competitive supervised approach: From-big-to-small (abbr. BTS) \citep{lee2019BTS}. The supervised method is trained on the Scenario dataset as we do. The training and validation datasets consist of 3600 and 900 images with resolution 320×320.

	\subsubsection{Quantitative comparison on the EndoSLAM dataset}
    	
    We perform the quantitative comparison on the EndoSLAM dataset, with the aim of demonstrating the feasibility and applicability of the proposed method.

	\textbf{Evaluation metrics.} As in previous works \citep{liu2020dense,shao2022AFM}, we adopt the standard evaluation metrics: root-mean-squared error (RMSE), the root-mean-squared logarithmic error (RMSE log), the absolute relative error (Abs Rel), the squared relative error (Sq Rel), and the accuracy (a1, a2, a3). 
	
	We use the median scaling to scale the predicted depth maps during the evaluation process \citep{tinghui2017median}. The scaled maps can be expressed as
	\begin{equation}
		{{\mathcal{\tilde{D}}}_{scaled}}=({\mathcal{\tilde{D}}}*(median(\mathcal{\widehat{D}})/median({\mathcal{\tilde{D}}})).
        \end{equation}

    Quantitative evaluation indicators for depth evaluation are as shown in \cref{res-table1}.
    \begin{table}[htb]
        \centering
        \caption{Quantitative evaluation indicators.}
        \includegraphics[width=0.85\linewidth]{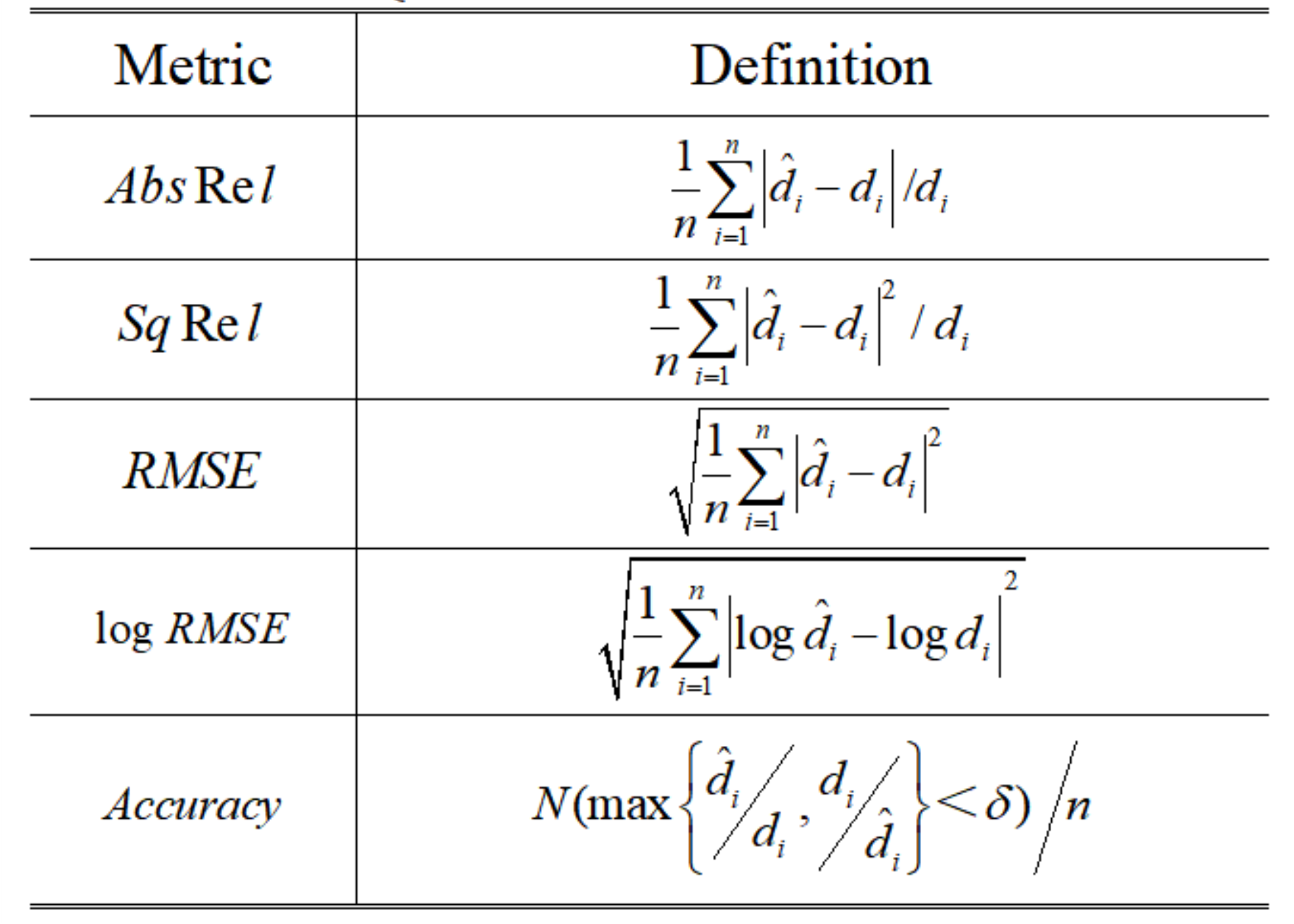}
        \label{res-table1}
    \end{table}
   
	\textbf{Ablation studies.} To better understand how the components of losses affect the overall performance, we conduct the four ablation study cases:
	\begin{itemize}    
        \item Case 1: depth and smoothness losses,
        \item Case 2: depth and smoothness losses, with gradient and normal losses,
        \item Case 3: depth and smoothness losses, with geometric loss,
        \item Case 4: depth and smoothness losses, with gradient, normal losses and geometric loss.
	\end{itemize}

 	\begin{figure*}[htb]
		\centering
		\includegraphics[width=0.95\linewidth]{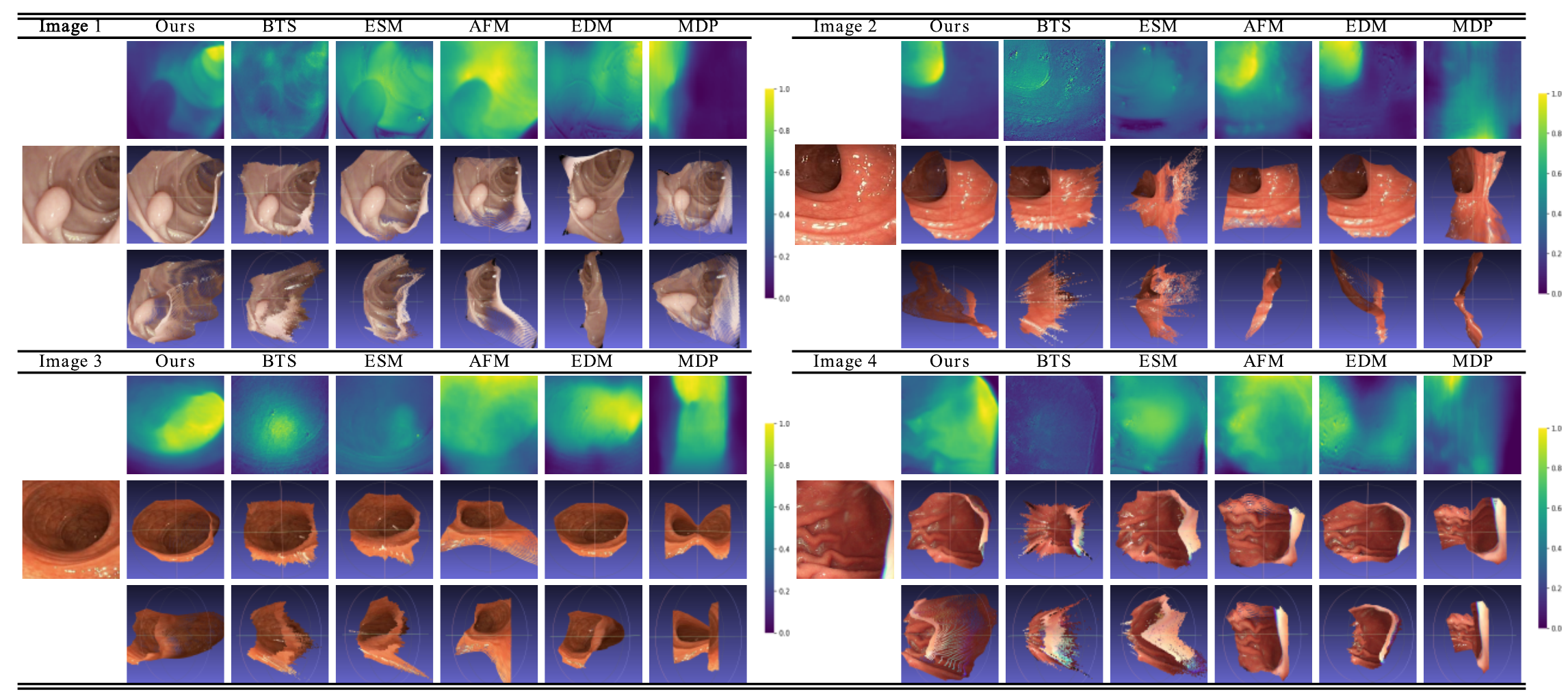}
		\caption{Qualitative results on the clinical images (single frame). Compared with our method, other methods show inaccurate depth predictions, such as missed polyps and edges, specular reflections, misinterpreted lumen geometry, distorted folds and fluctuations.}
		\label{fig9:expr-real-1}
	\end{figure*}
    
	As listed in \cref{table2:ablation}, we can observe that the gradient and normal losses reduce the RMSE and Abs Rel by 10\% and 20.4\% , respectively. The geometric loss reduces the RMSE and Abs Rel by 13\% and 25.4\%, respectively. Combing the gradient, normal and geometric losses reduce the RMSE and Abs Rel by 22\% and 36.3\%, respectively, compared to the depth and smoothness losses. This demonstrates the efficacy and advantages of our methods for dealing with depth estimation in endoscopic scenes. In order to compare and analyze contributions of each component, we visualize the estimated depth maps side by side, as shown in \cref{Visualization-ablation study} with several samples, it can be observed that the results in case2 are more sensitivity to edges and small structures due to the gradient and normal losses, but the relative distance relationship is not ideally. The estimated depth map in case 3 is accurate due to constraining the global geometric anatomy by the geometric consistency loss, but not as pronounced as in case 2 on edges and small structural local geometry regions. The results in case4 shows significant performance by correctly recovering stepped edges, small structures and global geometric surfaces structures.

	\textbf{Quantitative results.} As shown in \cref{table3:endoslam}, our method outperforms the compared methods. It is worth mentioning that while our method is trained only on the synthetic scenario dataset, it achieves mean RMSE values of 0.066, 0.029, and 0.139 for the stomach, small intestine, and colon frames, respectively. In particular, our predictions are consistent with camera light bursts and slight depth changes, which produce the least RMSE errors for all organs and demonstrate the method's adaptability across different organs. In addition, the severe reflections, illumination variations, and low-texture regions tend to result in large depth errors. Our method is robust to these edge cases, even in comparison to the ESM and AFM, which are especially designed to cope with illumination changes. Hence a much lower result on the Sq Rel is achieved.

	In \cref{fig7:expr-a}, a qualitative comparison is made with a more intuitive comparison of the depth prediction details, and the sensitivity to brightness variations. Our method is able to estimate the relatively far regions more accurately than the remaining ones, and the heatmaps also indicate that the errors significantly decrease for the pixels representing regions far from 0.6. Besides, it can be inferred that the biggest advantage of geometric consistency loss is to increase texture awareness whereas gradient and normal losses can preserve more detailed information. Their collaboration significantly improves the performance supported by lower RMSE and Sq Rel values. Particularly,  in  the colon images with specular reflections, we observe corresponding noise in the depth maps of BTS, ESM, EDM and MDP, but not for AFM and our method. Moreover, the error map results show that our method clearly detects the anatomical boundaries of the digestive tract and can provide consistent depth maps under large variations in the texture and lighting conditions.

	\subsubsection{Generalization to the ColonDepth dataset}

	The generalizability of the proposed method across datasets is verified on the ColonDepth dataset (described in \cref{table4:colondepth} and \cref{fig8:expr-b}). The superior performance demonstrates excellent generalizability, and the mean RMSE values on the ColonDepth dataset are 12.604 (T1-L1), 9.930 (T2-L2), and 13.893 (T3-L3). \begin{figure*}[htb]
		\centering
		\includegraphics[width=0.95\linewidth]{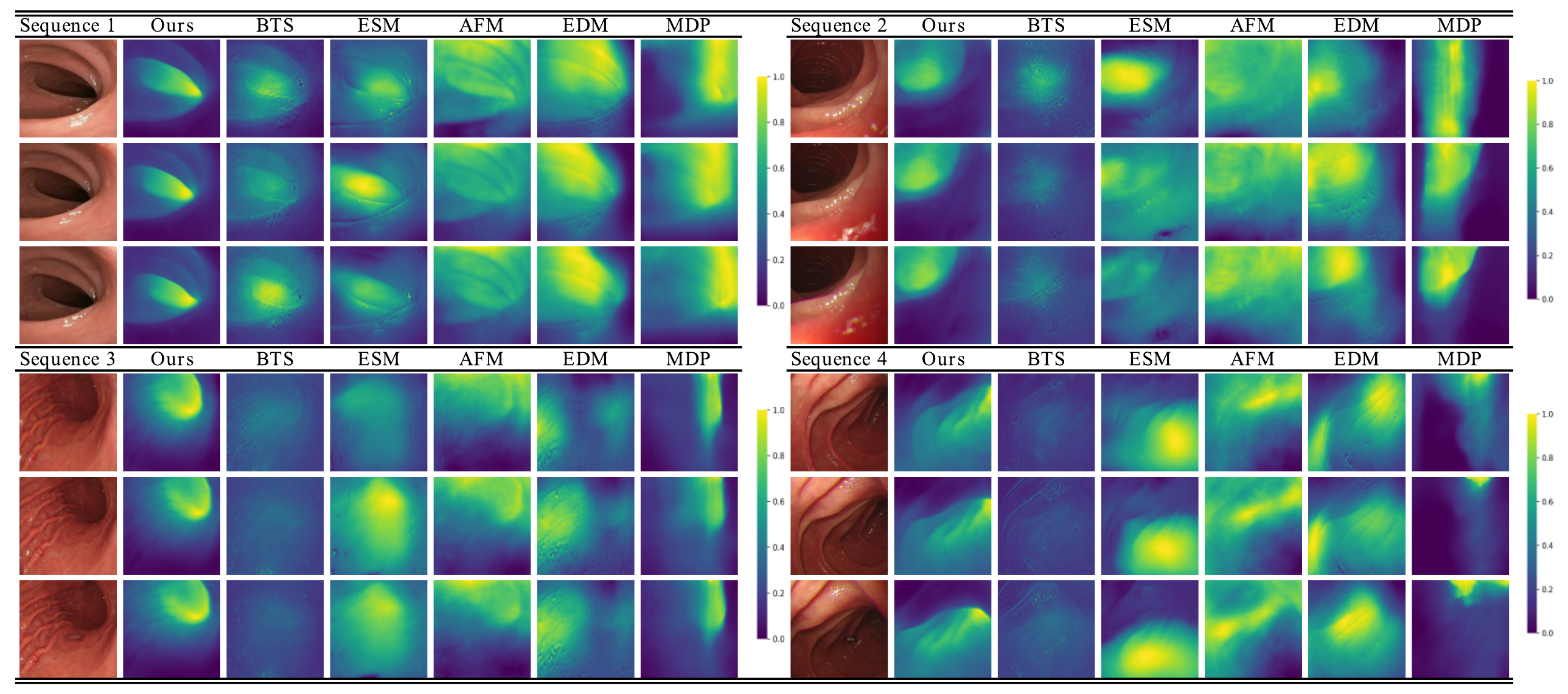}
		\caption{Qualitative results on the clinical images (consecutive frames). Compared with the results generated by other methods, our depth maps are smoother and consistent avoiding the influence of specular reflections and inconsistent lighting.}
		\label{fig10:expr-real-2}
	\end{figure*} Moreover, the visual results show that our method clearly detects the anatomical boundaries of the digestive tract and provides more continuous depth maps under varying texture and lighting conditions.

	\subsubsection{Validation on clinical images}
	
	We evaluate our method qualitatively on clinical images, as shown in \cref{fig9:expr-real-1}. The test images consist of polyps on Kvasir dataset (image 1), colon with specular reflections (image 2), duodenal lumen (image 3) and stomach folds (image 4). For visualization purposes, we transform each depth map into two 3D views based on the calibrated endoscope parameters. We respectively observe the missing polyps and edges in the results generated by the EDM and MDP methods in image 1, the obvious specular reflections of BTS, ESM and EDM in image 2, the lumen geometry misinterpreted of BTS, ESM, AFM and MDP in image 3, and the distorted folds and fluctuations of BTS, AFM and MDP in image 4. 
 
\begin{figure*}[htb]
    	\centering
    	\includegraphics[width=0.95\linewidth]{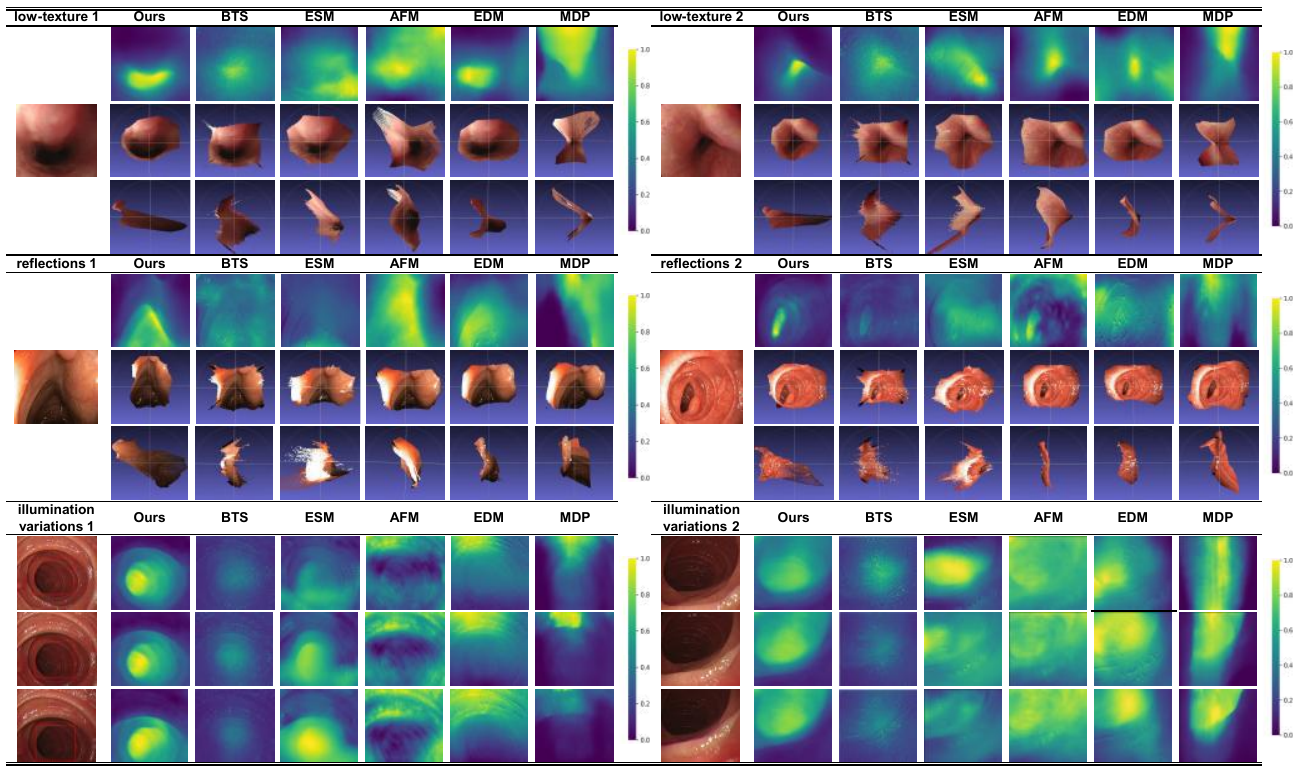}
        \caption{Three different types of scenarios and different challenges: low-texture regions, reflections, and illumination variations.}
        \label{RS:expr-real-3}
    \end{figure*}

    We also verify the depth estimation consistency for sequences as shown in \cref{fig10:expr-real-2}. The acquisition of clinical endoscopy videos is approved by the ethics committee of the partner hospital. The endoluminal sequences are disturbed by specular reflections and inconsistent lighting. The photometric inconsistency leads to depth estimation discontinuities between consecutive frames. In comparison, our depth maps are smoother in space and temporally consistent, thus avoiding fluctuations caused by specular reflections and inconsistent lighting. Due to the geometric consistency, the proposed approach achieves the most reasonable lumen structure details compared to other methods. In addition, we analyze the depth estimation ability for different types of scenes and different challenges, as shown in \cref{RS:expr-real-3}, providing depth maps and point clouds in low-texture areas, reflections and illumination changes, respectively. Particularly, in the low-texture areas, our method does not fail in low-texture images, while the depth maps obtained by other methods are not ideal. In four sets of images with reflections and illumination changes, BTS, ESM, AFM, and EDM are often disturbed by reflections and illumination variations. In contrast, our method can estimate the depth of stepped edges and small structures, and provide consistent depth maps under large variations in the texture and lighting conditions. Furthermore, our method also handles the demands for real-time depth estimation with an average inference time of 120 frames per second. Detailed results can be viewed in Supplementary Video.

	\subsubsection{Discussion}
 
    The proposed method is initially designed for and validated on interventional endoscopy data, and generalizes well across different datasets and clinical images. However, we are confident that it is also applicable to monocular endoscopy of other anatomies. We contend that the frequently-appearing inconsistent illumination between adjacent frames is one of the major difficulties for self-supervised methods, such as ESM, AFM, EDM and MDP, which are often disturbed by reflections and illumination variations. To a certain extent, the affine calibration (eg. ESM) and the appearance flow (eg. AFM), which are designed for capable of compensating brightness variations, add to the complexity of the task. Supervised methods, such as BTS and our method, use the depth difference between synthetic data and predicted data as a penalty term to bypass the appearance difference. Anatomical structures estimated results on the clinical images, such as image 1 , image2 , sequence 1 and sequence 2, show that the BTS and our method is superior to the self-supervised method. In conconst to BTS, our method takes into account the important geometric structural consistency, and the results of visual ablation experiments and different types of scenes can intuitively explain how the geometric consistency leads to more consistent depth maps.
 
    \section{Conclusion}
     In this paper, we perform a detailed analysis of a depth estimation network for endoscopic scenes, especially highlighting the challenges of mapping synthetic samples to real ones under low-texture images, reflections and illumination variations conditions. We propose a RGB-Depth scenario dataset to reduce the cost of learning surgical skills in simulation and real-world domains. In addition, we introduce a novel geometry-aware loss to allow a comprehensive geometric representation. In contrast to the previously proposed methods for depth estimation, our method gives the best quantitative validation outcomes on the EndoSLAM dataset, and generates more consistent depth maps and reasonable anatomical structures for clinical images. Finally, the outcomes demonstrate the strength and generalizability of our approach. 
     
     \textbf{Limitations and future work.} There are limitations of our method that need to be taken into consideration in future work. Our depth predictions suffers from scale ambiguity. In fact, the scale ambiguity is an intrinsic limitation that exists in the monocular depth estimation algorithms, such as Endo-SfM \citep{ozyoruk2021endoslam}, Endo-Depth-and-Motion \citep{Recasens2021EDM} and AF-SfM \citep{shao2022AFM}. The issue can be alleviated by increasing the possibilities of photometric model to recover the absolute scale. Depth estimation procedure can be regarded as an optimization problem, and multi-parameter optimization of photometric cost and regularization cost can be solve by representative computational intelligence algorithms, like monarch butterfly optimization (MBO) \citep{MBO-2015}, elephant herding optimization (EHO) \citep{EHO-2015}, earthworm optimization algorithm (EWA) \citep{EWA-2018}, moth search (MS) algorithm \citep{MS-2018}, Harris hawks optimization (HHO) \citep{HHO-2019}, Slime mould algorithm (SMA) \citep{SMA-2020}, hunger games search (HGS) \citep{HGS-2021}, Runge Kutta optimizer (RUN) \citep{RUN-2021}, colony predation algorithm (CPA) \citep{CPA-2021}, and differential evolution (DE) algorithm \citep{DE-2020,DE-2022}. In this way, we could recover the true scale of the endoscopy scenarios, which is what we currently do not have and will work on as a future direction.

    \section{Acknowledgements}
     This paper was supported by National Natural Science Foundation of China (U20A20195), the State Key Laboratory of Robotics (2023-Z08), Liaoning Province Science and Technology (2021JH2/10300058) and Collaborative Innovation Research of Medicine Engineering Combination of Shenyang (21-172-9-08, 20-205-4-015). 










\printcredits

\bibliographystyle{cas-model2-names}

\bibliography{cas-refs.bib}



\end{document}